\documentclass[12pt,reqno,a4paper]{amsart}
\usepackage{amsmath,amssymb,amsfonts,amscd}
\usepackage[mathscr]{eucal}
\usepackage{comment}
\usepackage{color}
\usepackage{hyperref}
\usepackage{float} 
\usepackage[pdftex]{graphicx}
\usepackage{cleveref}
\usepackage{subfig}

\usepackage{longtable}

\date{}
\title[Growing an architecture for a neural network]
{Growing an architecture for a neural network}
\author{Sergei~Khashin}
\address{Department of Mathematics,  Ivanovo State University,  Russia}
\email{khash2@gmail.com}
\author{Ekaterina~Shemyakova}
\address{Department of Mathematics \& Statistics,  University of Toledo, Ohio, USA}
\email{ekaterina.shemyakova@utoledo.edu}

\theoremstyle{definition}

\newtheorem{remark}{Remark}[section]

\newcommand{\x}{\mathbf{x}} 
\newcommand{\y}{\mathbf{y}} 
\newcommand{\w}{\mathbf{w}} 
\newcommand{\sign}{\mathop{\mathrm{sign}}}

\begin{document}

\begin{abstract}
 We propose a new kind of automatic architecture search algorithm.
The algorithm alternates pruning connections and adding neurons, and it is not restricted to layered architectures only.
Here architecture is an arbitrary oriented graph with some weights (along with some biases and an activation function), so there may be no layered structure in such a network. The algorithm minimizes the complexity of staying within a given error.   
We demonstrate our algorithm on the brightness prediction problem of the next point through the previous points on an image. Our second test problem is the approximation of the bivariate function defining the brightness of a black and white image. Our optimized networks significantly outperform the standard solution for neural network architectures in both cases. 
\end{abstract}
\maketitle
%
%
\section{Introduction}\label{sec.intro}

(Artificial) deep neural networks are nowadays very popular 
algorithms aiming to imitate processes inside a human brain. They train by examples and are shown to be very effective for pattern recognition problems of different kinds. 
Despite an avalanche of applications, much is unknown about neural networks from the mathematical point of view. Much is done by guessing and by whatever worked well in similar problems. 

One important open problem is   determining the best
architecture for a neural network. For a layered network this means to determine the
number of neurons and the number of layers for a neural network. 
Among the main approaches to   automatic architecture search are the following. See also survey~\cite{wistuba2019survey}.

\begin{enumerate}
 \item \emph{Empirical/statistical methods} that choose the weights according to the effect they make on the model’s performance, see, e.g. ~\cite{benardos2002prediction}. 
 \item \emph{Evolutionary algorithms} that
 start with selecting parent networks, then proceed with combination and mutations, and selecting the best ones. The algorithms then repeat by assigning the best ones as new parents. See   
e.g.~\cite{Neuro_symb_BENARDOS2007,Neuro_symb_Koza1991GeneticGO,ElsBAD}.
 \item  \emph{Pruning methods} that start with a larger than necessary multilayer network and then remove neurons that have little contribution to the solution. There are several different ways to decide which neuron is not needed, see e.g.~\cite{reed1993pruning,mozer1989skeletonization,Karnin1990,lecun1990optimal,castellano1997}. 
Pruning does not lead~\cite{segee1991fault} to the increase of fault tolerance of the system. Among known disadvantages is that one usually does not know a priori how large the original network should be. Also starting with a large network  could be excessively costly to trim the unnecessary units. 
\item  \emph{Constructive methods} that 
start with an initial network of a small size, and then incrementally add new hidden neurons and/or hidden layers until some prescribed error requirement is reached or if no performance improvement can be observed, see e.g. surveys~\cite{kwok1997constructive} (add a neuron, add a layer)
\cite{lee2012book}, and e.g. papers~\cite{MaKhorasani2003,Ash1989,weng1996constructive,prechelt1997,fahlman1990cascade,kwok1996bayesian,kwok1997objective,Shaw2019SqueezeNASFN}.
Among known disadvantages of these methods is that the size of the obtained multilayer networks are reasonable but rarely ``optimal''.
\item \emph{Cell-based methods} create the architecture from a smaller-sized blocks,  
see e.g.~\cite{zoph2018block,Shaw2019SqueezeNASFN,wu2019fbnet}. 
\end{enumerate}

Our \emph{growing architecture} algorithm
is a combination of ideas of \emph{both} prunning and constructing algorithms.  We also extend our domain to a more general one (that include layered networks as a particular case).
For us an architecture is an arbitrary oriented graph with some weights (along with some biases and an activation function), so there may be no layered structure in such a network. 
We compare our optimized network with the large number of networks with standard  architectures. We show that for the same error we can have a significantly smaller complexity. 



In recent years, we have seen the ever-increasing efficiency of neural networks. At the same time, their complexity is growing. Here we measure the \emph{complexity of a neural network} by the number of \emph{weights}, the values of which are selected in the learning process. 
Those who do not work directly with neural networks usually expect the complexity of the network to be tens, hundreds, at most thousands. In reality, the complexity of modern neural networks is much higher. Thus, for the standard MNIST handwritten digit classification problem, the number of learnable parameters in the best networks is hundreds of thousands and millions, while there are only $60,000$ training examples 
(small square $28 \times 28$ pixel grayscale images of handwritten single digits between $0$ and $9$).

Minimization of the network complexity is the goal of our work. 


Most of our computations are realized in C++
instead of some conventional package (e.g. TensorFlow/Python). This is because layered architectures are the main objective of such specialized packages, and dealing with non-layered ones presents such difficulties that outweigh their conveniences. 

The structure of the paper is as follows. In~\cref{sec:prelim}, we introduce our notation. In~\cref{sec:algorithm}, we describe our ``architecture growing'' algorithm. In~\cref{sec:comparison}, we present the results of our experiments with the brightness prediction problem and compare our algorithm with polynomial regression and standard neural networks. In~\cref{sec:conclusions}, we summarize our results.

\section{Preliminaries. Architecture and complexity}
\label{sec:prelim}
Let  $f(\x,\w)$ be a function that represents a feedforward neural network, where $\x$ be an input vector and $\w$ be a vector of learnable parameters (weights). 
For regression tasks we minimize 
 \emph{target function}
\begin{equation} \label{eq:S}
   S = \sum (f(\x_i,\w) - \y_i)^2 \ .
\end{equation}
We do not consider \emph{convolutional neural networks}: 
all training vectors $\x_i$ have a fixed length. The training dataset is represented in the form of a matrix $A$, each line of which first contains the value $\y_i$, and then coordinates of the vector $\x_i$.

\emph{Fully connected} are layered networks where each neuron receives the values of all neurons from the previous layer. 
Our algorithm allows networks of even more general type, where every neuron receives all input values and also values from all previous neurons. 
We shall call such \emph{maximally fully connected}. By changing weights, one can include layered fully and not fully connected networks into the network of such a general type.
Each neuron is a function of ${\Bbb R}^n \rightarrow {\Bbb R}$
of the form
\[
 (z_1, \dots z_n) \mapsto g(w_0 + w_1 z_1 + \dots w_n z_n)
\]
where $(z_1, \dots z_n)$ are neuron's input arguments, $(w_0, w_1,\dots, w_n)$ are the corresponding weights and $g: {\Bbb R} \rightarrow
{\Bbb R}$ is some \emph{activation function}.
In this paper we consider some most common ones, see details in~\cref{subsec:activation}. 

{Hardware specification: Intel(R) Pentium(R) CPU G4500 @ 3.50GHz, 32G byte mem. and Intel(R) Core(TM) i7-8565U CPU @ 1.80GHz 16 G byte mem.}

{Software specification: Visual Studio 2019 Community (C++), Python 3.9.5, PyCharm 2019 Community Edition, Numpy 1.19.5, TensorFlow 2.5.0.}

\section{Architecture growing algorithm}
\label{sec:algorithm}
The rough idea is to remove redundant connections in the neural network, while possible, and then to add a new neuron to its beginning, with running training processes in between. Then again remove some connections, while possible, and when it not possible, add a neuron, and so on.
We first describe elements of the algorithm and then put it all together at the end of the section. 

\subsection{Connection removal procedure}
\label{subsec:redundant}
\begin{enumerate}
\item Find three connections with the least (w.r.t. their absolute value) weights.
\item Create three different networks, by removing one of these three connections in each case. In every case start the learning process for the resulting network to minimize~\eqref{eq:S} within the specified time $\Delta$. 
\item Choose from the three reduced optimized networks the one
with the smallest error and optimize it with training time $3 \Delta$.
\item While removing a connection, it may turn out that this connection is the only input connection for some
 neuron, i.e. it works by the formula
\[
 F(x_1) = f(w_0 + w_1 x_1) \ .
\]
In this case, we remove such a neuron, and approximate its action by linear function
\[
G(x_1) = v_0 + v_1 x_1 \, ,
\]
where we choose $(v_0, v_1)$ s.t. to minimize the deviation
\[
 || G-F ||^2 = \int_{t_0}^{t_1} (F(t)-G(t))^2 dt \, ,
\]
where $[t_0, t_1]$ is the interval of values taken by the parameter $x_1$
for the whole training matrix.
\item While removing a connection, it may turn out that the value of some neuron does not participate anymore in further
calculations. We remove such a neuron.

\end{enumerate}
Repeat the ``one connection removal procedure'' until the error increases by no more than 
$(1+ \varepsilon)$ times, where
$\varepsilon$ is small enough. In our experiments $\varepsilon = 0.006$.
\begin{remark}
 One might expect that by removing connections in a neural network, we increase its error. However, as we discovered, it is frequently not the case if the original network was not trained well enough. In such cases the quality of the ``reduced'' network can be much improved after the optimization. Our experiments showed that in the beginning, when the connections are started to be removed, the error in the network 
almost does not grow until we reach a ``saturation'' point where any attempt to remove any further connection results in a noticeable increase of the error. 
The boundary $1.006$ was chosen based on the results of these experiments.
\end{remark}
\subsection{Adding a neuron procedure} 
\label{subsec:extraneuron}

In the course of the algorithm, we run the connection removal procedure until the error increase is too large. After that we do the following.

\begin{enumerate}
 \item Add one extra neuron to the very beginning of the neural network and connect it 
with all input parameters and with all other
neurons of the original network.
 \item Set all the weights of all new connections equal to $0$. Thus, the computations in this new neural network go exactly by the same algorithm as in the original one, and the error will remain the same. 
 \item Retrain the new network with training time $5\Delta$. The error of the network decreases.
\end{enumerate}

\subsection{Architecture growing algorithm}
We start with an arbitrary architecture and then execute the following procedure. 
\begin{enumerate}
\item Remove all redundant connections as described in~\cref{subsec:redundant}.
\item If the complexity of the network reaches the preset limit, end the procedure.
\item Add a neuron as described in~\cref{subsec:extraneuron}.
\item Return to step (1).
\end{enumerate}

\begin{figure}[htbp]
\center{
  \includegraphics[width=0.79\textwidth]{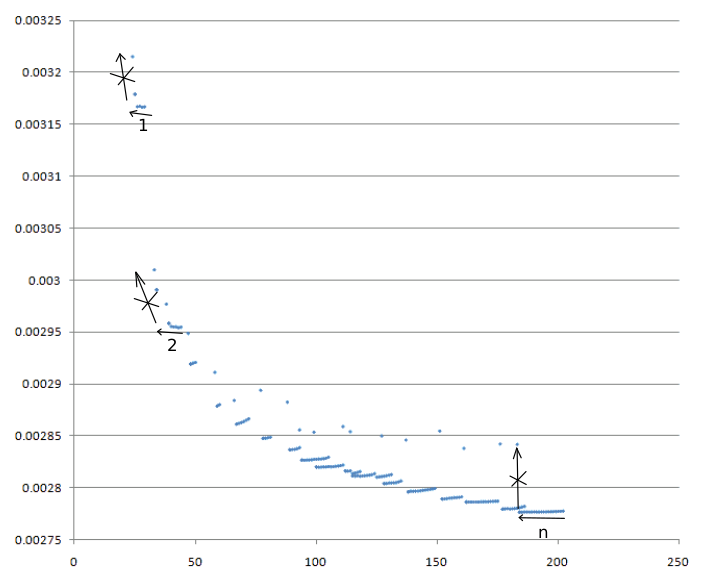}}
\vskip0.1cm
\caption{Dependence of the error on the complexity in the course of running our algorithm.}\label{fig:errorcomplexity1}
\end{figure}

\begin{figure}[htbp]
\center{
   \includegraphics[width=0.79\textwidth]{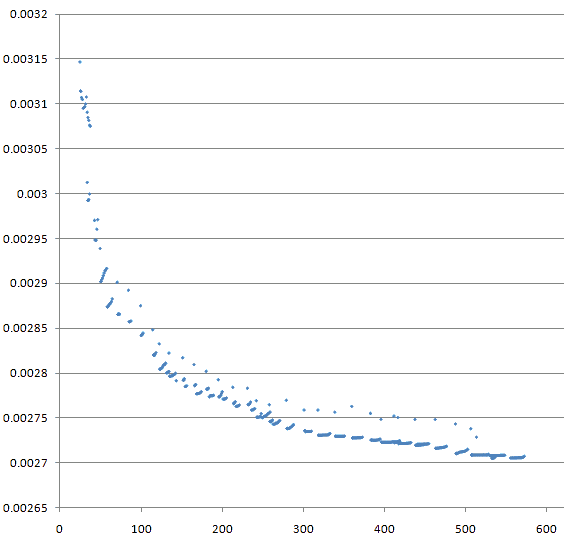}
}
\vskip0.1cm
\caption{Dependence of the error (vertical axis) on the complexity (horizontal axis).}\label{fig:errorcomplexity2}
\end{figure}

On \cref{fig:errorcomplexity1} and~\cref{fig:errorcomplexity2} are two typical graphs of the  dependence of the error on the complexity  that we obtain executing our algorithm. 
On \cref{fig:errorcomplexity1}, the starting point of the algorithm is represented by the blue point above the start of $\xleftarrow[1]{}$.  Removing connections, we move in the direction indicated by
$\xleftarrow[1]{}$, from the right to the left: complexity decreases while the error slightly increases. At some moment, removing a connection leads to a sudden jump in error (see along vertical-ish arrow). 

Once this happens we stop removing connections and add one more neuron. This moves us to the start of $\xleftarrow[2]{}$. From there we continue the process of removing connections (and thus move in the direction indicated by $\xleftarrow[2]{}$) until we reach the jump in error, and then the process repeats. 
\section{Comparison with other approaches on brightness prediction problem} 
\label{sec:comparison}
To illustrate our idea, consider the brightness prediction problem for  an image point knowing the brightness of several previous points.

\subsection{Neuro network built using our approach} 
The data is kept in a table, where rows and columns are $x$ and $y$ coordinates of the point and values are 
the numbers from $0$ to $255$. 

The \emph{previous points} are ordered as shown in \cref{table:orderingpoints}. Here $Y$ is the current point, and the first column and the first row give 
the $(x,y)$-coordinates of the points relative to $Y$.  The other numbers in the table indicate the order in which the points will be considered.
 
 \begin{table}[htbp]
\centering
\begin{tabular}{c|ccccccc}\hline
        & -3  & -2  & -1  & 0  & 1  & 2  & 3  \\ \hline
    -3  &     &     & 15  & 13 & 16 &    &    \\
    -2  &     & 10  &  7  &  5 &  8 & 11 &    \\
    -1  &  14 &  6  &  2  &  1 &  3 &  9 & 17 \\
     0  &  12 &  4  &  0  &  Y &  - &  - & - \\ \hline
  \end{tabular}
\caption{Ordering of the previous points. Here $Y$ is the current point.}
\label{table:orderingpoints}
\end{table}
  To normalize the numbers, we subtract from all values of $Y$ the brightness of the previous (left) point and divide all values by $100$. So building the forecast by $5$ points, there will be only $4$ input parameters in the neural network. For the experiments we choose a graphics file of size $512 \times 512$. For $4$ previous points, the resulting training file looks as in \cref{table:trainingfile}.
  \begin{table}[htbp]
\centering
\begin{verbatim}
                            y,   x0,   x1,   x2,   x3
                            -0.01,-0.01, 0.00, 0.00, 0.00
                            0.01, 0.01, 0.00,-0.04, 0.01
                            -0.05,-0.05, 0.00, 0.01,-0.01
                            …
                            0.11,-0.01,-0.09, 0.08,-0.03
                            0.05,-0.03,-0.12, 0.00,-0.11
                            0.06,-0.05,-0.08, 0.14,-0.05
\end{verbatim}
\caption{An excerpt from the training file for $4$ previous points.}
\label{table:trainingfile}
\end{table}
We present and compare the results of our experiments
in \cref{table:long} and~\cref{table:best_errors}.
  \newpage
\subsection{For comparison: linear and polynomial approximation}

As a starting point, we compare the performance of our optimized networks with the results obtained using linear/polynomial regressions:
\[
 \y_i \approx P_d (\w,\x_i) \ ,
\]
where $P_d$ is a polynomial in $\x_i$ of degree $d$, and $\w$ is the vector of weights. \cref{table:errors_poly} contains the mean square errors (over all points of the image) 
in such approximation with polynomials of degrees $1$, $2$, and $3$.

\begin{table}[htbp]
\centering
\begin{tabular}{|c|c|c|c|c|c|c|}\hline
  & \multicolumn{2}{c|}{$\deg=1$} & \multicolumn{2}{c|}{$\deg=2$} & \multicolumn{2}{c|}{$\deg=3$}  \\    
  \hline 
  number of points    & Error & Complexity & Error  &  Complexity   & Error   & Complexity  \\
  \hline
    3  & 0.00466 & 3 & 0.00465 & 6   & 0.00462  & 10  \\
     4  & 0.00417 & 4 & 0.00416 &10 & 0.00381 &20  \\
    5  & 0.00402  & 5 & 0.00399 & 15 & 0.00354 & 35 \\
    6  & 0.00366 &  6 & 0.00366 &21 & 0.00359 & 56 \\
     8  & 0.00359 & 8 & 0.00359 &36 & 0.00350 &120 \\
   10  & 0.00336 &10 & 0.00325 &55 & 0.00274 &220 \\
    12  & 0.00333 &12 & 0.00322 &78 & 0.00269 &364\\
    18  &  0.00331 &18 & 0.00318 &171 & ? &1140  \\ \hline
  \end{tabular}
\caption{Errors in polynomial regressions depending on the number of previous points. The complexity is the number of parameters.}
\label{table:errors_poly}
\end{table}
Here the number of points used for the 
approximation is larger by $1$ than the number of input parameters of the network. The complexity of the network is the number of parameters. For example, the cubic approximation of the brightness of a point by the previous $10$ points has $220$ parameters
(that are coefficients of the corresponding polynomial) and gives error $S=0.00274$. 

To see the error in the original units and before normalization, one takes values of errors $S$ from  \cref{table:errors_poly} and calculate the value of  $100 \cdot \sqrt{S}$. For example, for  $S=0.00274$, it is approximately $5.23$. 

Comparing these with the numbers in \cref{table:long} and~\ref{table:best_errors}, once sees
that, as expected, polynomial regressions work with much less efficiency. 

\subsection{For comparison: neural networks with some fixed architectures}
\label{subsec:activation}
Here we compare our optimized networks with the networks having some standard architectures. Specifically, we consider a large number of 3-layered networks and maximally fully connected networks. 
The considered 3-layered networks have from $6$ to $20$ neurons in the hidden layers, and the considered maximally
fully connected networks have up to $500$--$600$ neurons.

\begin{figure}[htbp] 
    \centering
        \includegraphics[width=\textwidth]{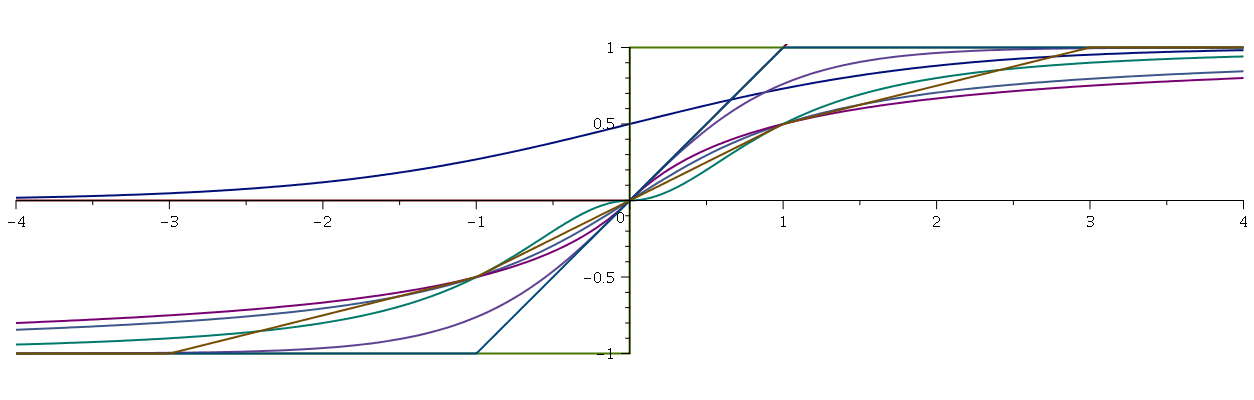}%
        \label{fig:456}
\caption{Graphs of activation functions tested.}        
\end{figure}

The choice of the activation function is a part of the architecture. We consider the most popular ones, and through many computations, choose the one that is the most efficient. The following activation functions were considered
(see their graphs on \cref{fig:456}).
\begin{align*}
f_0(x) &= x \\
f_1(x) &= \left\{
  \begin{array}{ll}
    0      \, ,    & x<0 \\
    x      \, ,  & x \geq 0   
  \end{array}
\right.  \ \text{(ReLU)} \\
f_2(x) &= \frac{1}{1+e^{-x}} \ \text{(sigmoid)}
\\
 f_3(x) &= \left\{
  \begin{array}{ll}
    -1      \, ,    & x<0 \\
    +1      \, ,  & x \geq 0   
  \end{array}
\right. \ \text{(sign)} \\
 f_4(x) &= 2 \, \frac{\arctan(x)}{\pi} \\
 f_5(x) &= \frac{x}{1+|x|} \\
 f_6(x) &= \sign(x) \frac{x^2}{1+x^2} \\
 f_7(x) &= \tanh(x) \ \text{(hyperbolic tangent)} \\
 f_8(x) &= \left\{
  \begin{array}{ll}
    x          & |x|\le 1 \\
    \sign(x)    & |x| > 1  
  \end{array}
\right. \\
f_9(x) &= \left\{
  \begin{array}{ll}     
    x/2          & 0< x\le 1 \\
    (x+1)/4      & 1 < x < 3  \\
    1            & 3 \ge x   
  \end{array}
\right. \ \text{and then extended using the fact that it is odd.}
\end{align*}

All but the first three functions are odd. 
Function $f_6$ was found to be the most effective. 
        
\begin{figure}[htbp]
\center{
  \includegraphics[width=0.72\textwidth]{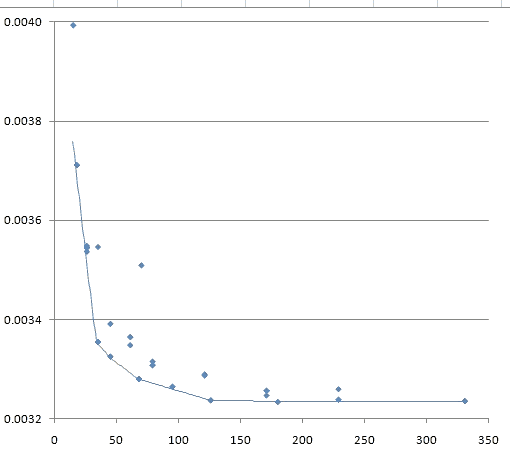}
}
\caption{Complexity and achieved quality (error) for standard networks. The piece-wise linear envelope interpolates the discrete data to give an estimation for 
the best quality with a given complexity and for the best complexity with a given quality.}
\label{fig:comparison_otherneuro}
\end{figure}

Note that since the minimization process of the neural network is stochastic, repeating the experiment several times, we obtain different results. Thus, for the same architecture, we repeat the experiment $10$ times and choose the best result. In \cref{fig:comparison_otherneuro} are the results of the performance on $5$ input points by standard networks.

The envelope from below (the piece-wise linear line) in \cref{fig:comparison_otherneuro} gives interpolation of our data and allows us to approximate the smallest expected error for a given complexity, or vice versa, the lowest 
complexity for a given error. We use this to compare best achievable results for networks of different complexities. The approach in particular makes sense since the results are of stochastic nature. 

We obtained similar results through a large number of experiments with $4$ to $12$ of the input points for $3$-layer and full-connected networks.

\subsection{Our algorithm}

On the same training matrices, we optimize the architectures using our algorithm.

\cref{table:long} 
allows us to compare the lowest achievable complexities (for our and the standard networks). For example, in the last line: with $11$ input points, among our experiments there is an optimized architecture network with a complexity $128$. The corresponding error is $0.00251157$. Then the best complexity $312$ of the standard networks with the same error
is obtained based on the envelope shown in \cref{fig:comparison_otherneuro}. We see that for the same error, the complexity of the optimized network is significantly less than that of the standard network.

\begin{table}[htbp]
\centering
\begin{tabular}{|c|c|c|c|}\hline
  Number  & Error   & Best complexity of  & Complexity of  \\ 
  of points &  & standard networks for this error  & optimized networks  \\ \hline
  4  & 0.00358090 &  23 &  21 \\
       & 0.00345647 &  29 &  22 \\
       & 0.00329252 &  61 &  50 \\
       & 0.00324596 & 114 & 100 \\
       & 0.00323453 & 179 & 119 \\ \hline
    5  & 0.00323960 &  22 &  18 \\
       & 0.00290990 &  52 &  50 \\
       & 0.00282196 & 329 & 100 \\ \hline
    6  & 0.00316931 &  27 &  21\\
       & 0.00286380 &  56 &  50\\
       & 0.00277786 & 180 & 100\\
       & 0.00276770 & 359 & 118\\ \hline
    7  & 0.00268228 & 200 & 103\\
       & 0.00266749 & 300 & 116\\
       & 0.00266493 & 374 & 123\\ \hline
    8  & 0.00266364 & 100 &  66\\
       & 0.00259725 & 162 & 104\\ \hline
   10  & 0.00263869 & 100 &  67\\
       & 0.00258670 & 150 &  94\\
       & 0.00254739 & 205 & 123\\ \hline
   11  & 0.00265986 & 100 &  52\\
       & 0.00256666 & 200 &  92\\
       & 0.00251747 & 300 & 121\\
       & 0.00251157 & 312 & 128\\ \hline
  \end{tabular}
\caption{Comparison of the lowest achieved complexities by our approach  and by standard (for the same 
error).}
\label{table:long}
\end{table}

A different type of comparison is given in \cref{table:best_errors}.
For a given number of points, it shows
the best achieved quality (i.e. minimizing $S$) by standard and our optimized networks.
\begin{table}[htbp]
\centering
\begin{tabular}{|c|c|c|c|c|}\hline
   & \multicolumn{2}{c|}{standard networks} &  \multicolumn{2}{c|}{our optimized networks}  \\
  \hline 
  number of & smallest achieved  & corresponding & smallest achieved & corresponding  \\
  previous points & error  &  complexity & error &  complexity \\ \hline
   5  &   0.00323453  &  180  & 0.003245946  &  78  \\
   6  &   0.00282148  &  346  & 0.002768079  & 415  \\
   7  &   0.00276767  &  361  & 0.002721608  & 339  \\
   8  &   0.00266489  &  376  & 0.002628534  & 261  \\
   9  &   0.00257263  &  277  & 0.00259757   & 101  \\
   11 &   0.00252952  &  232  & 0.002547398  & 124  \\
   12 &   0.00251157  &  313  & 0.002490456  & 173  \\  \hline
  \end{tabular}
\caption{The best results (as measured by the best quality, that is the smallest error) that are achieved in our experiments, for standard and optimized networks.}
\label{table:best_errors}
\end{table}


The constructed by our algorithm network architectures can be represented as a graph: neurons correspond to vertices, and connections between neurons correspond to graph edges.
For better visualization the input values and edges outgoing from these vertices are not shown. 
Below are a few obtained graphs in ``circular'' form where vertices of graph are located in the vertices of a regular polygon. 
\begin{figure}[htbp] 
    \centering
     \includegraphics[width=0.5\textwidth]{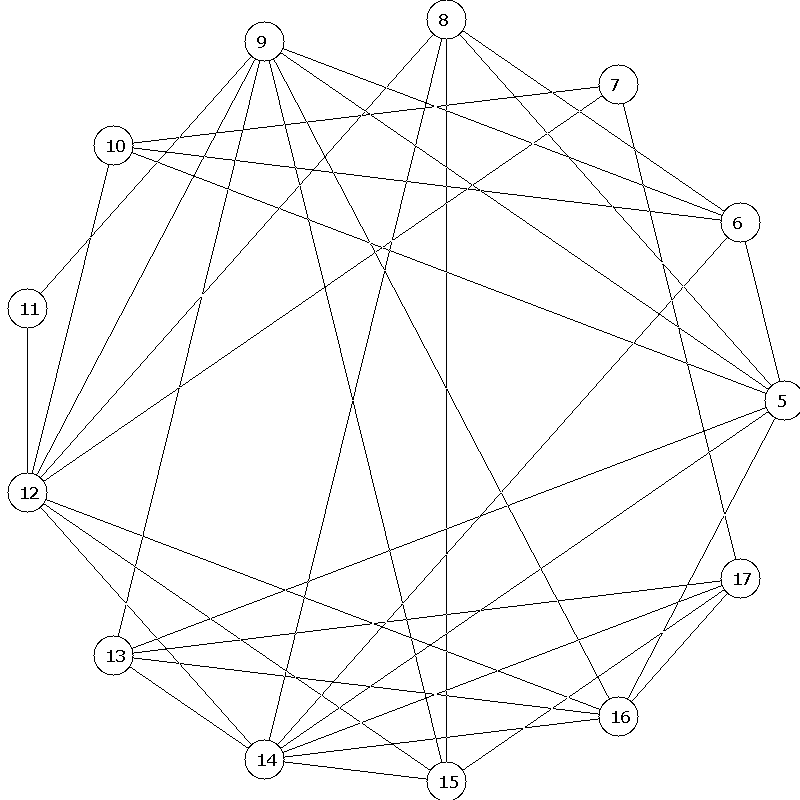}%
        \label{fig:NW77}
\caption{$77$ weights}        
\end{figure}

\begin{figure}[htbp] 
    \centering
    \includegraphics[width=0.5\textwidth]{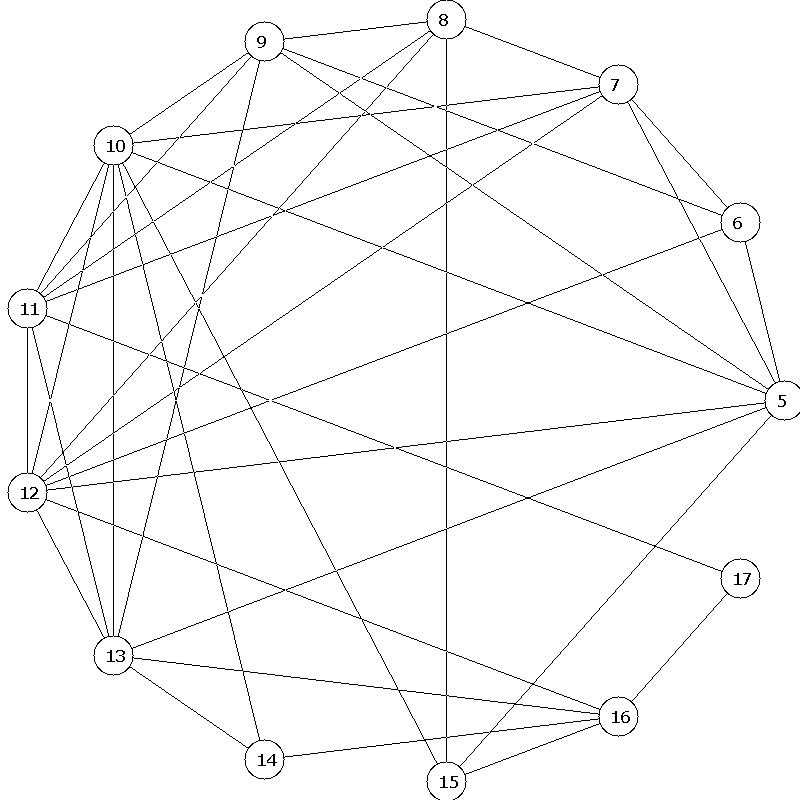}%
        \label{fig:Nw120}%
     \caption{$120$ weights}
\end{figure}
\begin{figure}[htbp] 
    \centering
         \includegraphics[width=0.5\textwidth]{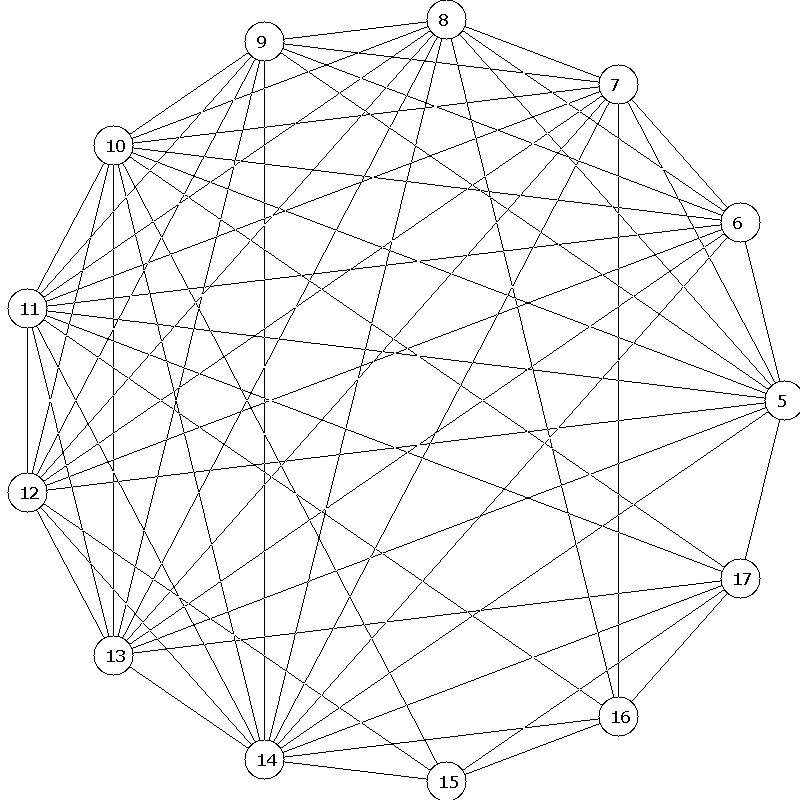}%
        \label{fig:NW112}  
        \caption{$112$ weights}
\end{figure}
\begin{figure}[htbp] 
    \centering
        \includegraphics[width=0.5\textwidth]{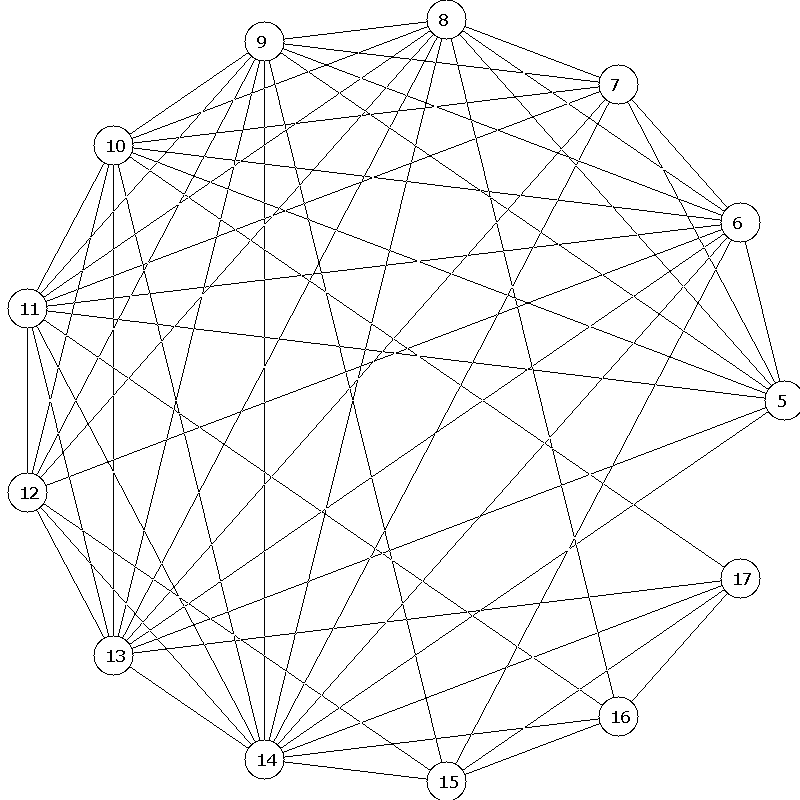}%
        \label{fig:Nw184}%
                \caption{$184$ weights}
\end{figure}

\section{Comparison with other approaches on image approximation problem}

\begin{figure}[htbp] 
    \centering
     \includegraphics[width=\textwidth]{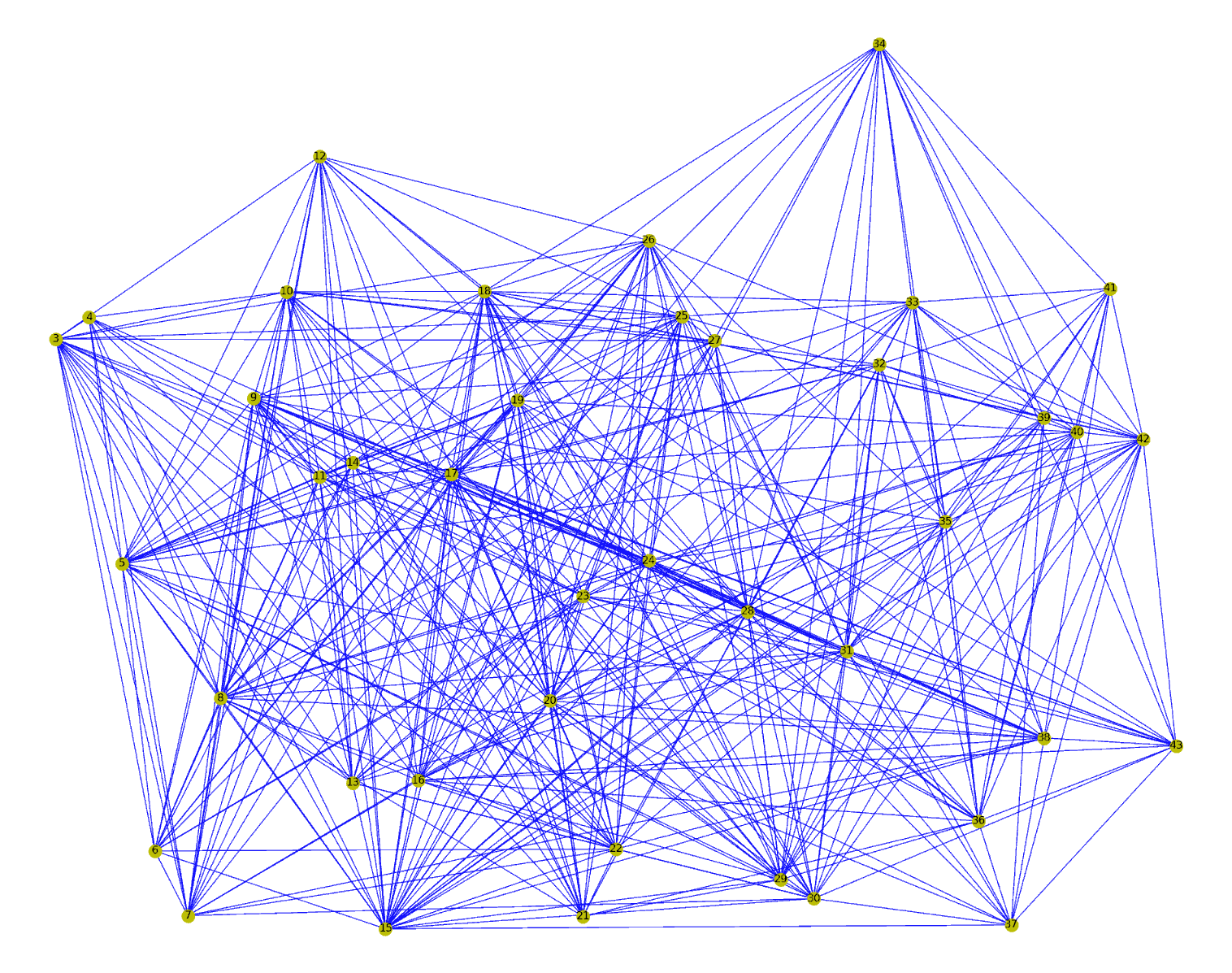}
     \caption{The graph of one of the optimized networks. Specific parameters are as follows: $43$ neurons (including $2$ input neurons), complexity is $503$, error $0.004607364$. Arrows go from right to left, from vertices labeled with large numbers to vertices labeled with smaller numbers.}
      \label{fig:res2graph_optimized}
\end{figure}

Here we look into approximating of a  black and white graphics file of size $512 \times 512$ by  a function of two input variables, $f(x,y)$. We consider 3-layered architectures with two inputs, $N_1$ 
neurons in the first layer, $N_2$ neurons in the second layer,
and $1$ output neuron. Here $4 \leq N_1, N_2 \leq 40$. 
We run the learning process on \emph{TensorFlow/Keras}. We optimize these networks using our approach.  
 On \cref{fig:res2graph_optimized} is the graph of one of the optimized by us networks. 

\begin{figure}[htbp] 
    \centering
 \includegraphics[width=0.9\textwidth]{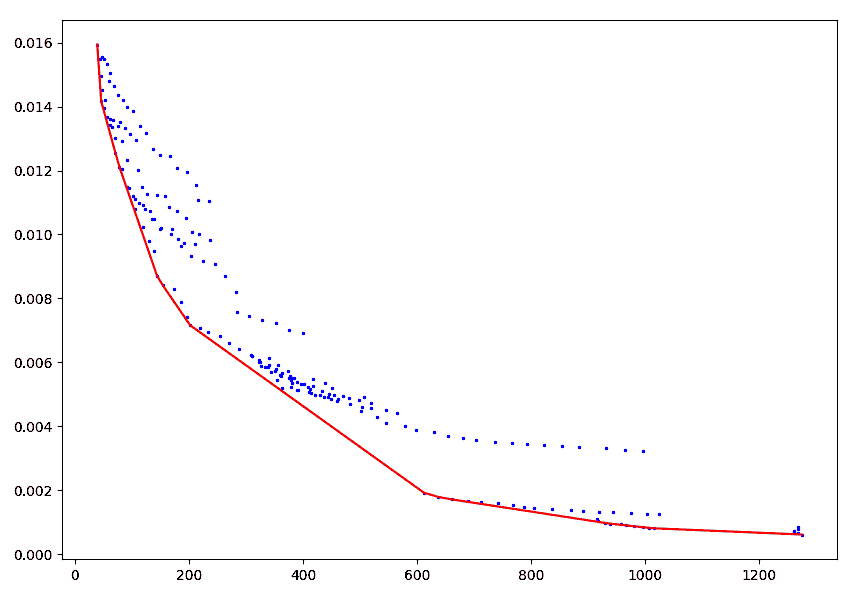}
     \caption{Blue points are the obtained complexities/errors by the standard networks. The red curve is the envelope of this data extrapolating the obtained results and allowing the comparison for others complexities/errors.}
      \label{fig:res2_envelope_standard_network}
\end{figure}

A comparison of the described networks one can see in \cref{table:final_comparison_table_for_2ndexperiment}. 
To compare complexities we use the same approach as in Sec.~\ref{sec:comparison}, i.e. construct an envelope from  
the below to extrapolate discrete data that we get from computations, see \cref{fig:res2_envelope_standard_network}. One can see that for the same error, our optimized networks offer significantly smaller complexities.

 \begin{table}[htbp]
\centering
\begin{tabular}{|ccc|}\hline
     Given    & Best complexity  & Best complexity \\
     error    & of standard architectures  & of our optimized networks   \\
      \hline 
    0.014 &  67 & 48   \\
    0.013 &  87 & 63   \\
    0.011 &  117 & 99   \\
    0.010 &  123 & 118   \\
    0.009 &  145 & 138   \\
    0.008 &  182 & 169   \\
    0.007 &  237 & 214   \\ 
    0.006 &  308 & 292   \\ 
    0.005 &  408 & 371  \\ 
    0.0045 & 474 & 410   \\
    0.0040 &  590 & 449   \\ 
    0.0035 &  768 & 488   \\ 
    0.0030 &  987 & 527  \\ 
    0.00225 & 1765 & 562   \\ 
    0.0020  & --- & 606 \\
    0.0012  & --- & 847 \\
    0.0010  & --- & 921 \\
    0.0007  & --- & 1158 \\
    0.0006138 & --- & 1274   \\ 
    \hline
  \end{tabular}
\caption{Comparing best complexities achieving the given errors.}
\label{table:final_comparison_table_for_2ndexperiment}
\end{table}

\section{Conclusions}
\label{sec:conclusions}

We propose a new kind of automatic architecture search algorithm.
The algorithm  alternates pruning connections and adding neurons, and it does not restrict itself to layered networks only. Instead,
we search for an architecture among arbitrary oriented graph with some weights (along with some biases and an activation function), so there may be no layered structure in such a network. The goal is to minimize the complexity staying within a given error.  

We begin with any standard architecture and create our optimized one by pruning and letting it grow, pruning and again letting it grow, and so on.

For large networks, where the number of connections counts in hundreds, the complexity of the optimized network is
$2-2.5$ times smaller  than that of a standard network with the same error.
For small networks, where the number of connections counts in tens, the complexity decreases by a factor of $1.25-2$. Here by standard networks, we mean the best results obtained with $3$-layer and full-connection networks.
 
The algorithm can be sped up by for example not  considering every connection while pruning. 
%

 \bibliographystyle{plain}

\begin{thebibliography}{10}

\bibitem{Ash1989}
T.~Ash.
\newblock Dynamic node creation in backpropagation networks.
\newblock {\em Connection Science}, 1(4):365--375, 1989.

\bibitem{zoph2018block}
Vijay B.~Zoph, Vasudevan, Jonathon Shlens, and Quoc~V Le.
\newblock Learning transferable architectures for scalable image recognition.
\newblock In {\em Proceedings of the IEEE conference on computer vision and
  pattern recognition}, pages 8697--8710, 2018.

\bibitem{Neuro_symb_BENARDOS2007}
P.G. Benardos and G.-C. Vosniakos.
\newblock Optimizing feedforward artificial neural network architecture.
\newblock {\em Engineering Applications of Artificial Intelligence},
  20(3):365--382, 2007.

\bibitem{benardos2002prediction}
PG~Benardos and G~Cl Vosniakos.
\newblock Prediction of surface roughness in cnc face milling using neural
  networks and taguchi's design of experiments.
\newblock {\em Robotics and Computer-Integrated Manufacturing},
  18(5-6):343--354, 2002.

\bibitem{castellano1997}
Giovanna Castellano, Anna~Maria Fanelli, and Marcello Pelillo.
\newblock An iterative pruning algorithm for feedforward neural networks.
\newblock {\em IEEE transactions on Neural networks}, 8(3):519--531, 1997.

\bibitem{Karnin1990}
E.D. Karnin.
\newblock A simple procedure for pruning back-propagation trained neural
  networks.
\newblock {\em IEEE Transactions on Neural Networks}, 1(2):239--242, 1990.

\bibitem{Neuro_symb_Koza1991GeneticGO}
J.~Koza and J.~F. Rice.
\newblock Genetic generation of both the weights and architecture for a neural
  network.
\newblock In {\em IJCNN-91-Seattle International Joint Conference on Neural
  Networks}, volume~2, pages 397--404, 1991.

\bibitem{kwok1997objective}
Tin-Yan Kwok and Dit-Yan Yeung.
\newblock Objective functions for training new hidden units in constructive
  neural networks.
\newblock {\em IEEE Transactions on neural networks}, 8(5):1131--1148, 1997.

\bibitem{kwok1996bayesian}
Tin-Yau Kwok and Dit-Yan Yeung.
\newblock Bayesian regularization in constructive neural networks.
\newblock In {\em International Conference on Artificial Neural Networks},
  pages 557--562. Springer, 1996.

\bibitem{kwok1997constructive}
Tin-Yau Kwok and Dit-Yan Yeung.
\newblock Constructive algorithms for structure learning in feedforward neural
  networks for regression problems.
\newblock {\em IEEE transactions on neural networks}, 8(3):630--645, 1997.

\bibitem{lecun1990optimal}
Yann LeCun, John~S Denker, and Sara~A Solla.
\newblock Optimal brain damage.
\newblock In {\em Advances in neural information processing systems}, pages
  598--605, 1990.

\bibitem{lee2012book}
Tsu-Chang Lee.
\newblock {\em Structure level adaptation for artificial neural networks},
  volume 133.
\newblock Springer Science \& Business Media, 2012.

\bibitem{wistuba2019survey}
A.~Rawat M.~Wistuba and T.~Pedapati.
\newblock A survey on neural architecture search.
\newblock {\em arXiv:1905.01392 [cs.LG]}, 2019.

\bibitem{MaKhorasani2003}
L.~Ma and K.~Khorasani.
\newblock A new strategy for adaptively constructing multilayer feedforward
  neural networks.
\newblock {\em Neurocomputing}, 51:361--385, 2003.

\bibitem{ElsBAD}
R.~Miikkulainen, J.~Liang, E.~Meyerson, A.~Rawal, D.~Fink, O.~Francon, B.~Raju,
  H.~Shahrzad, A.~Navruzyan, N.~Duffy, and B.~Hodjat.
\newblock Ch.15 - evolving deep neural networks.
\newblock In {\em Artificial Intelligence in the Age of Neural Networks and
  Brain Computing}, pages 293--312. 2019.

\bibitem{mozer1989skeletonization}
Michael~C Mozer and Paul Smolensky.
\newblock Skeletonization: A technique for trimming the fat from a network via
  relevance assessment.
\newblock In {\em Advances in neural information processing systems}, pages
  107--115, 1989.

\bibitem{prechelt1997}
Lutz Prechelt.
\newblock Investigation of the cascor family of learning algorithms.
\newblock {\em Neural Networks}, 10(5):885--896, 1997.

\bibitem{reed1993pruning}
Russell Reed.
\newblock Pruning algorithms -- a survey.
\newblock {\em IEEE transactions on Neural Networks}, 4(5):740--747, 1993.

\bibitem{fahlman1990cascade}
C.~Lebiere S.E.~Fahlman.
\newblock The cascade-correlation learning architecture.
\newblock {\em Advances in Neural Information Processing Systems}, 2, 1990.

\bibitem{segee1991fault}
Bruce~E Segee and Michael~J Carter.
\newblock Fault tolerance of pruned multilayer networks.
\newblock In {\em IJCNN-91-Seattle International Joint Conference on Neural
  Networks}, volume~2, pages 447--452. IEEE, 1991.

\bibitem{Shaw2019SqueezeNASFN}
A.E. Shaw, D.~Hunter, F.N. Iandola, and S.~Sidhu.
\newblock Squeezenas: Fast neural architecture search for faster semantic
  segmentation.
\newblock {\em 2019 IEEE/CVF International Conference on Computer Vision
  Workshop (ICCVW)}, pages 2014--2024, 2019.

\bibitem{weng1996constructive}
Wei Weng and Khashayar Khorasani.
\newblock An adaptive structure neural networks with application to eeg
  automatic seizure detection.
\newblock {\em Neural Networks}, 9(7):1223--1240, 1996.

\bibitem{wu2019fbnet}
Bichen Wu, Xiaoliang Dai, Peizhao Zhang, Yanghan Wang, Fei Sun, Yiming Wu,
  Yuandong Tian, Peter Vajda, Yangqing Jia, and Kurt Keutzer.
\newblock Fbnet: Hardware-aware efficient convnet design via differentiable
  neural architecture search.
\newblock In {\em Proceedings of the IEEE/CVF Conference on Computer Vision and
  Pattern Recognition}, pages 10734--10742, 2019.

\end{thebibliography}

\end{document}